\documentclass{article}

\usepackage{arxiv}

\usepackage[utf8]{inputenc} 
\usepackage[T1]{fontenc}    
\usepackage{hyperref}       
\usepackage{url}            
\usepackage{booktabs}       
\usepackage{amsfonts}       
\usepackage{nicefrac}       
\usepackage{microtype}      
\usepackage{lipsum}
\usepackage{graphicx}
\usepackage{algorithmic}
\usepackage[tight,footnotesize]{subfigure}
\usepackage{caption}
\usepackage{adjustbox}
\usepackage{makecell} 
\usepackage[linesnumbered,ruled]{algorithm2e}

\usepackage{pifont}
\newcommand{\cmark}{\ding{51}}%
\newcommand{\xmark}{\ding{55}}%

\graphicspath{ {./images/} }

\title{Proof-of-TBI --- Fine-Tuned Vision Language Model Consortium and OpenAI-o3 Reasoning LLM-Based Medical Diagnosis Support System for Mild Traumatic Brain Injury (TBI) Prediction}

\author{
Eranga Bandara \\
Old Dominion University \\
Norfolk, VA, USA \\
\texttt{cmedawer@odu.edu} \\
\And
Ross Gore \\
Old Dominion University \\
Norfolk, VA, USA \\
\texttt{rgore@odu.edu} \\
\And
Sachin Shetty \\
Old Dominion University \\
Norfolk, VA, USA \\
\texttt{sshetty@odu.edu} \\
\And
Alberto E. Musto \\
Old Dominion University \\
Norfolk, VA, USA \\
\texttt{mustoae@odu.edu} \\
\And
Pratip Rana \\
Old Dominion University \\
Norfolk, VA, USA \\
\texttt{prana@odu.edu} \\
\And
Ambrosio Valencia-Romero \\
Old Dominion University \\
Norfolk, VA, USA \\
\texttt{a1valenc@odu.edu} \\
\And
Christopher Rhea \\
Old Dominion University \\
Norfolk, VA, USA \\
\texttt{crhea@odu.edu} \\
\And
Lobat Tayebi \\
Old Dominion University \\
Norfolk, VA, USA \\
\texttt{ltayebi@odu.edu} \\
\And
Heather Richter \\
Old Dominion University \\
Norfolk, VA, USA \\
\texttt{hrichter@odu.edu} \\
\And
Atmaram Yarlagadda \\
McDonald Army Health Center \\
Newport News, VA, USA \\
\texttt{atmaram.yarlagadda.civ@health.mil} \\
\And
Donna Edmonds \\
BRAINBox Solutions \\
Richmond, VA, USA \\
\texttt{dedmonds@brainboxinc.com} \\
\And
Steven Wallace \\
BRAINBox Solutions \\
Richmond, VA, USA \\
\texttt{swallace@brainboxinc.com} \\
\And
Donna Broshek \\
UVA Health \\
Charlottesville, VA, USA \\
\texttt{DKB6V@uvahealth.org} \\
}

\begin{document}
\maketitle
\begin{abstract}
Mild Traumatic Brain Injury (TBI) detection presents significant challenges due to the subtle and often ambiguous presentation of symptoms in medical imaging, making accurate diagnosis a complex task. To address these challenges, we propose Proof-of-TBI, a medical diagnosis support system that integrates multiple fine-tuned vision-language models with the OpenAI-o3 reasoning large language model (LLM). Our approach fine-tunes multiple vision-language models using a labeled dataset of TBI MRI scans, training them to diagnose TBI symptoms effectively. The predictions from these models are aggregated through a consensus-based decision-making process. The system evaluates the predictions from all fine-tuned vision language models using the OpenAI-o3 reasoning LLM, a model that has demonstrated remarkable reasoning performance, to produce the most accurate final diagnosis. The LLM Agents orchestrates interactions between the vision-language models and the reasoning LLM, managing the final decision-making process with transparency, reliability, and automation. This end-to-end decision-making workflow combines the vision-language model consortium with the OpenAI-o3 reasoning LLM, enabled by custom prompt engineering by the LLM agents. The fine-tuning of each vision-language model was conducted using the Unsloth library on Google Colab's Tesla GPUs. To ensure optimal performance on consumer-grade hardware, Low-Rank Adapters with 4-bit quantization (QLoRA) were employed. The prototype for the proposed platform was developed in collaboration with the U.S. Army Medical Research team in Newport News, Virginia, incorporating five fine-tuned vision-language models. The results demonstrate the transformative potential of combining fine-tuned vision-language model inputs with the OpenAI-o3 reasoning LLM to create a robust, secure, and highly accurate diagnostic system for mild TBI prediction. To the best of our knowledge, this research represents the first application of fine-tuned vision-language models integrated with a reasoning LLM for TBI prediction tasks.
\end{abstract}


\section{Introduction}

Mild TBI is a significant global health concern, affecting millions of individuals annually~\cite{midl-tbi-definision}. Despite its prevalence, the accurate diagnosis of mild TBI remains a formidable challenge. The subtle and often ambiguous nature of mild TBI symptoms, combined with variations in patient presentation, frequently leads to misdiagnosis or delayed detection. Traditional diagnostic methods, such as manual analysis of Magnetic Resonance Imaging (MRI) scans, are time-consuming and highly reliant on the expertise of clinicians, further complicating the process. These challenges underscore the need for innovative diagnostic systems that can improve the accuracy, reliability, and efficiency of TBI detection~\cite{mild-tbi-challenges}.

In recent years, advancements in Artificial Intelligence (AI) and machine learning have shown immense potential in the field of medical imaging, with vision-language models emerging as a transformative tool~\cite{tbi-ml-elsevior, vision-language-model}. Vision-language models combine image analysis and natural language processing, enabling them to interpret visual data and generate meaningful insights. These models, such as Meta's Llama-Vision, Pixtral, and Qwen2-VL~\cite{vistion-language-model-comparison, pixtral}, are pre-trained on vast datasets and are capable of understanding complex patterns in visual data. By fine-tuning these models with domain-specific datasets, such as labeled MRI scans, they can be adapted to address specialized tasks like the diagnosis of mild TBI. However, leveraging a single vision-language model for diagnosis has inherent limitations, including model biases and the potential for inconsistent predictions~\cite{vlm-image-classification}. To overcome these challenges, we propose ``Proof-of-TBI", a novel medical diagnosis support system that integrates a consortium of fine-tuned vision-language models with OpenAI-o3 reasoning LLM~\cite{o3, reasoning-llms}. In this study, we fine-tuned multiple vision-language models using a labeled dataset of TBI MRI scans, enabling them to effectively identify and diagnose TBI symptoms. To ensure efficient deployment on consumer-grade hardware, the models were optimized using quantized Low-Rank Adapters (QLoRA)~\cite{qlora}. The predictions from the ensemble of vision-language models were aggregated through a consensus-based decision-making process, facilitated by the OpenAI-o3 reasoning model, a specialized LLM designed for advanced reasoning tasks. The OpenAI-o3 model evaluates the predictions from all vision-language models in the consortium and determines the most accurate final diagnosis. This end-to-end decision-making process for TBI is fully automated, with LLM agents orchestrating the interactions between the vision-language model consortium and the OpenAI-o3 reasoning model, ensuring accuracy, efficiency, and transparency.

The paper presents the methodology and implementation of the ``Proof-of-TBI" platform, developed in collaboration with the U.S. Army Medical Research team in Newport News, Virginia. By integrating vision-language model consortium with OpenAI-o3 reasoning LLM, we aim to address the challenges of accurate mild TBI diagnosis and demonstrate the transformative potential of this innovative approach in the field of medical diagnostics. Following are our main contribution of this paper.

\begin{enumerate}
    \item Utilizing a fine-tuned vision-language model consortium to analyze MRI scans and identify TBI symptoms.
    \item Incorporating the OpenAI-o3 reasoning model to provide the final diagnosis based on the vision-language model consortium's predictions.
    \item Automating the end-to-end decision-making process for TBI using the vision-language model consortium and the OpenAI-o3 model, facilitated by LLM agents.
    \item Implementing the prototype of Proof-of-TBI, integrating three vision-language models, in collaboration with the U.S. Naval Medical Research Team in Norfolk, Virginia, USA.
\end{enumerate}

\begin{figure}[t]
\centering{}
\includegraphics[width=4.5in]{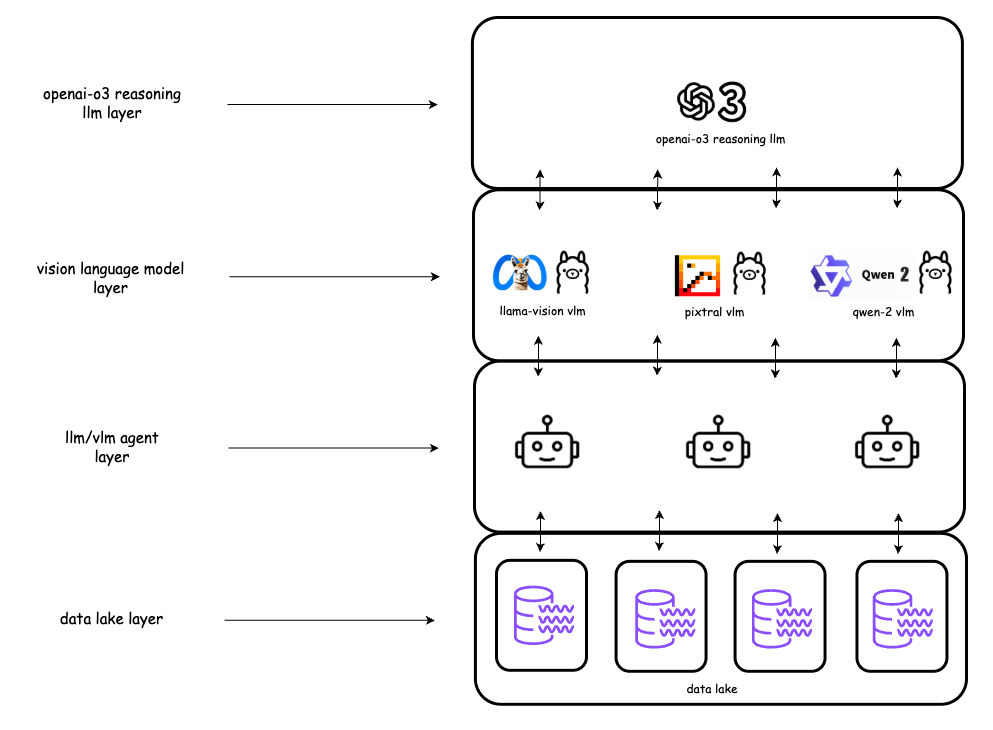}
\vspace{-0.1in}
\DeclareGraphicsExtensions.
\caption{Proof-of-TBI platform layered architecture.}
\label{indy528-architecture}
\end{figure}

\section{System Architecture}

Figure~\ref{indy528-architecture} describes the architecture of the platform. The proposed platform is composed of 4 layers: 1) Data Lake layer, 2) Blockchain/Smart Contract Layer, 3) LLM Layer, and 4) Data Provenance Layer, below is a brief description of each layer.

\subsection{Data Lake Layer}

The Data Lake layer of the platform serves as a centralized infrastructure for managing and storing the extensive dataset of MRI images required for training and fine-tuning vision-language models (VLMs) for Mild TBI diagnosis. This layer hosts a diverse collection of large-scale labeled datasets of MRI scans, encompassing various stages and symptoms of TBI. These labeled datasets are essential for developing the models' ability to accurately analyze and diagnose TBI symptoms from medical imaging~\cite{vision-language-model}.

Beyond simple storage, the Data Lake layer is designed to facilitate efficient data management and accessibility, ensuring that the high volume and variety of MRI data required for training and inference are seamlessly handled. Before being utilized for training the VLMs, the raw MRI data undergoes a series of pre-processing steps, including cleaning, anonymization, structuring, and normalization, to optimize its quality and readiness for effective learning. This layer also supports advanced querying and retrieval mechanisms, allowing researchers to analyze historical data and perform continuous model fine-tuning, ensuring the platform remains adaptive and effective in addressing the complexities of TBI diagnosis.

\subsection{LLM Agent Layer}

The LLM Agent Layer serves as the automation and orchestration core of the platform, enabling seamless integration and coordination among the various layers. This layer handles all custom prompt engineering required to facilitate interactions between the vision-language models (VLMs) and the OpenAI-o3 reasoning LLM. The LLM agents are responsible for prompting the fine-tuned vision-language models with MRI images to obtain their diagnostic predictions. These predictions are then aggregated and structured into a custom prompt, which is passed to the OpenAI-o3 reasoning LLM. The OpenAI-o3 model processes this input and provides the final diagnosis, leveraging its reasoning capabilities to synthesize and refine the information from the VLM consortium. By dynamically generating and managing prompts tailored to each model's input requirements, the LLM Agent Layer ensures effective communication between the vision-language models and the reasoning LLM. This orchestration not only enhances diagnostic accuracy but also automates the end-to-end decision-making process, ensuring the platform operates efficiently and reliably.

\subsection{Vision Language Model Layer}

The Vision-Language Model (VLM) layer forms the analytical core of the platform, powering its ability to analyze MRI scans and provide accurate diagnostic insights for Mild TBI. This layer comprises a consortium of fine-tuned vision-language models, each trained using labeled datasets of TBI MRI scans. These models are specialized to predict TBI-related observations from MRI scans, enabling the platform to provide precise and consistent medical diagnoses~\cite{vistion-language-model-comparison}.

The fine-tuned models operate on Ollama~\cite{ollama}, a framework optimized for efficient deployment and management of large language models (LLMs) and vision-language models. This ensures that the platform can handle complex inference tasks while maintaining optimal performance on consumer-grade hardware. As shown in Figure~\ref{llama2-flow}, the platform’s LLM Agent seamlessly interact with the vision-language models through Ollama's API, LlamaIndex, and LangChain~\cite{llamaindex, langchain}. These integrations facilitate efficient model orchestration, data retrieval, and result aggregation.

\begin{figure}[t]
\centering{}
\includegraphics[width=4.5in]{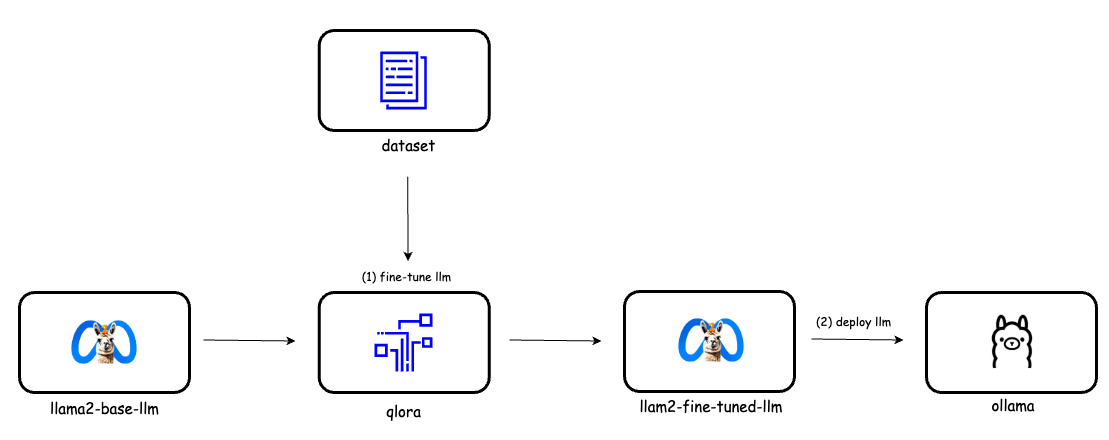}
\vspace{-0.1in}
\DeclareGraphicsExtensions.
\caption{Fine-tune Vision LLMs with Qlora and deploy with Ollama.}
\label{llm-fine-tune}
\end{figure}

\begin{figure}[t]
\centering{}
\includegraphics[width=4.5in]{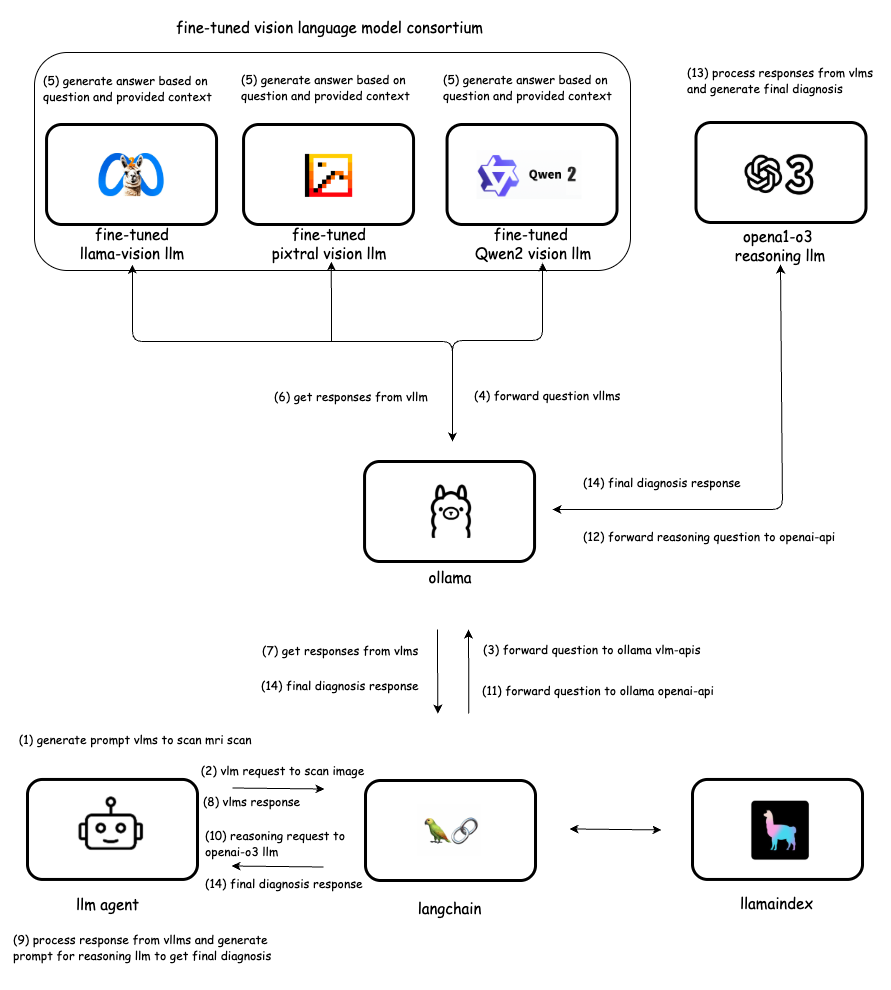}
\vspace{-0.1in}
\DeclareGraphicsExtensions.
\caption{Vision LLM integration flow with Ollama LLM-API, LlamaIndex, LangChain and Smart Contracts.}
\label{llama2-flow}
\end{figure}

\subsection{Reasoning LLM Layer}

The Reasoning LLM Layer represents the advanced reasoning capabilities of the platform, leveraging first-generation reasoning model OpenAI o3~\cite{o3}. OpenAI o3 model demonstrated remarkable performance in handling complex reasoning tasks, serving as a foundational step to deliver more robust and refined reasoning capabilities. In the platform, the final prediction of TBI diagnosis is performed by the OpenAI-o3 reasoning LLM. Predictions generated by the consortium of vision-language models (VLMs) are collected and evaluated by OpenAI-o3 to identify the most consistent and reliable diagnostic outcomes. This reasoning LLM synthesizes the diverse inputs from the VLM consortium, ensuring that the final diagnosis is accurate, consistent, and well-validated. The LLM Agent Layer manages this interaction with OpenAI-o3 by constructing custom prompts based on the VLM outputs and handling the overall decision-making process. This approach guarantees transparency and trustworthiness in the diagnostic outcomes, making the OpenAI-o3 Reasoning LLM Layer a crucial component of the platform’s success in predicting mild TBI.

\section{Platform Functionality}

There are four main functionalities of the platform: 1) Data Lake Setup, 2) Vision language model Fine-Tuning, 3) TBI Prediction by Fine-tuned VLMs, and 4) Final TBI Diagnosis Prediction by OpenAI-o3 reasoning LLM. This section goes into the specifics of these functions.

\subsection{Datalake setup}

The Data Lake serves as the foundational layer of the platform, designed to store and manage the extensive dataset of labeled MRI scans required for fine-tuning vision-language models. The Data Lake primarily contains MRI scans from patients diagnosed with Mild TBI, along with their corresponding textual observations. These observations include detailed annotations, such as the stage of TBI, specific symptoms, and other diagnostic information provided by medical experts. By consolidating this data into a centralized repository, the Data Lake ensures seamless access and efficient management of high-quality training data.

This dataset is pivotal for fine-tuning the vision-language models, enabling them to learn complex patterns in MRI images and accurately associate them with textual descriptions of TBI symptoms~\cite{tbi-mdpi}. Additionally, the Data Lake is optimized for scalability and retrieval, facilitating advanced querying and preprocessing of the data. Before being used for fine-tuning, the raw MRI scans and associated textual data undergo preprocessing steps such as anonymization, normalization, and structuring to ensure readiness and quality. By providing a robust and efficient infrastructure for data management, the Data Lake plays a crucial role in the platform’s ability to develop specialized models for accurate and reliable TBI diagnosis.

\subsection{Vision Language Model Fine-Tuning}

The next critical step in the platform's workflow is fine-tuning the vision-language models using the labeled MRI scan data stored in the Data Lake. This step involves training multiple vision-language models, including Llama-Vision~\cite{vistion-language-model-comparison}, Pixtral~\cite{pixtral}, and Qwen2-VL~\cite{qwen2}, with the labeled dataset to adapt these models to the specific domain of Mild TBI diagnosis. Fine-tuning is conducted using the Unsloth library, which provides the tools necessary for large-scale model adaptation~\cite{llamafactory-unsloth}. 

Through this process, the general-purpose vision-language models are transformed into specialized models capable of identifying TBI-related features and symptoms from MRI scans. This fine-tuning step enables the models to associate complex visual patterns in MRI data with textual observations, resulting in a powerful, domain-adapted diagnostic tool for TBI. 

To optimize the process, the fine-tuning integrates Quantized Low-Rank Adapters (LoRA)~\cite{qlora}, incorporating 4-bit quantization. This approach ensures that the fine-tuned models can operate efficiently on consumer-grade hardware without sacrificing performance, as depicted in Figure~\ref{llm-fine-tune}. Once fine-tuned, these quantized models run seamlessly on Ollama, a deployment framework optimized for vision-language model operation.

\subsection{TBI Prediction by Fine-tuned VLMs}

Once the vision-language models are fine-tuned, the next critical functionality of the platform is to predict TBI diagnoses using the fine-tuned VLM consortium. When a new MRI scan is received, the platform's LLM Agent interacts with the fine-tuned vision-language models through the Ollama API~\cite{ollama} to extract diagnostic insights. 

To facilitate this interaction, the LLM Agent employs custom prompt engineering to inject relevant context and the MRI scan data into tailored prompts for the vision-language models. These prompts are designed to maximize the diagnostic accuracy of the models by providing them with precise contextual information. The fine-tuned models then analyze the MRI scan, identify potential TBI-related symptoms, and generate predictions~\cite{vlm-image-classification}. These diagnostic predictions are sent back to the LLM Agent, which collects and organizes the outputs for further reasoning. This step ensures that the fine-tuned models’ specialized capabilities are leveraged to provide detailed and reliable diagnostic insights into TBI-related conditions.

\subsection{Final TBI Diagnosis Prediction}

To ensure accuracy and reliability, the platform employs a consensus-based decision-making mechanism for final TBI diagnosis. Instead of relying on the prediction from a single vision-language model, the platform aggregates responses from multiple fine-tuned models in the consortium. These responses are then evaluated and synthesized by the OpenAI-o3 reasoning LLM to determine the most consistent and reliable diagnosis. This consensus-driven approach forms the foundation of the Proof-of-TBI system, significantly enhancing the robustness of diagnostic outcomes~\cite{reasoning-llms}.

As a specialized reasoning LLM, OpenAI-o3~\cite{o3} is uniquely capable of analyzing and comparing the outputs from the vision-language models. It integrates these insights to produce a final diagnosis that is both precise and trustworthy. To instruct the OpenAI-o3 model effectively, the LLM Agent generates custom prompts by organizing and embedding the responses from the vision-language models. These prompts, as illustrated in Figure 1, provide the reasoning model with structured inputs, enabling it to evaluate the predictions and finalize the diagnosis with accuracy. This integration of consensus-based prediction and advanced reasoning ensures the platform delivers a high level of transparency, reliability, and diagnostic precision.

This consortium-based approach enhances diagnostic accuracy by mitigating the limitations of individual models, leveraging the collective intelligence of multiple fine-tuned vision-language models. By orchestrating this process transparently and securely through LLM Agent, the platform ensures robust and reliable TBI diagnosis. This step highlights the transformative potential of integrating consensus-driven AI to improve diagnostic decision-making for mild TBI.

\begin{figure}
\centering{}
\includegraphics[width=4.5in]{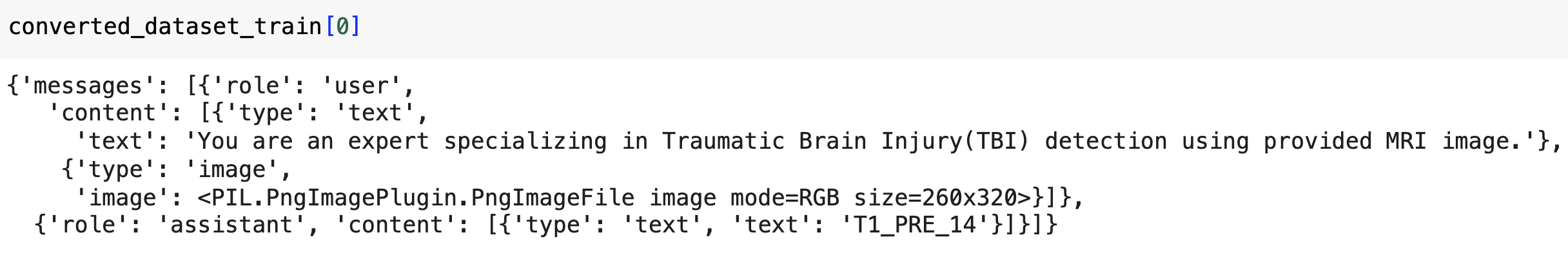}
\vspace{-0.1in}
\DeclareGraphicsExtensions.
\caption{The required data format of the unsloth library to fine-tune the vision language model.}
\label{unsloth-format}
\end{figure}

\begin{figure}
\centering{}
\includegraphics[width=4.5in]{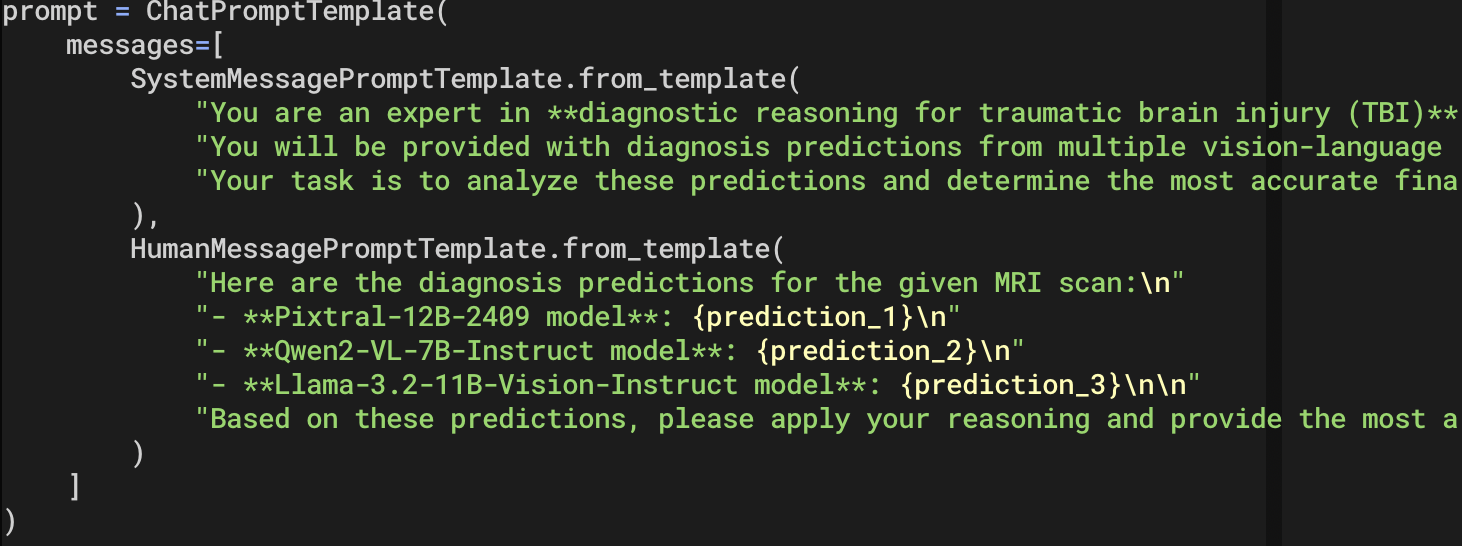}
\DeclareGraphicsExtensions.
\caption{Prompt for OpenAI-o3 reasoning LLM for final diagnosis reasoning.}
\label{prompt}
\end{figure}

\section{Implementation and Evaluation}

The implementation of the Proof-of-TBI platform was conducted in collaboration with the U.S. Army Medical Research team in Newport News, Virginia, USA. This partnership facilitated access to critical domain expertise and datasets necessary for building an advanced diagnostic system for Mild TBI. The core analytical layer comprises three fine-tuned vision-language models, including Llama-Vision, Pixtral and Qwen2-VL~\cite{llama-3, vistion-language-model-comparison}. These models were fine-tuned using a combination of publicly available datasets from platforms like Kaggle and Hugging Face, as well as confidential datasets provided under strict ethical and privacy guidelines. The datasets included labeled MRI scans along with detailed annotations of TBI-related symptoms, which were instrumental in adapting the models to the specific domain of TBI diagnosis. Fine-tuning was performed using the Unsloth library on Google Colab, leveraging Tesla TPUs for efficient training. The Unsloth library requires the fine-tuning dataset to be in a conversation format~\cite{llamafactory-unsloth}. To meet this requirement, we converted the original dataset into the required format, as illustrated in Figure~\ref{unsloth-format}. This format includes key fields such as `content` (which serves as the instruction to the LLM), `type` (specifying the image content), and `instruction` (providing contextual guidance to the LLM). After fine-tuning, the models were quantized using QLoRA~\cite{qlora}, a process that enables efficient operation on consumer-grade hardware. This optimization was critical for deploying the fine-tuned models on Ollama, a framework designed for lightweight yet high-performance execution of vision-language models. The OpenAI-o3 Reasoning LLM also running with the Ollama. 

The LLM Agent interacts with the vision-language models and OpenAI-o3 reasoning LLM through the Ollama API, leveraging LangChain and Llama-Index for efficient query handling and response management~\cite{langchain, llamaindex}. Based on the predictions of the VLMs, OpenAI-o3 LLM diagnosis the final TBI prediction. As illustrated in Figure~\ref{prompt}, custom prompts~\cite{prompt-engineering} are utilized to instruct the OpenAI-o3 Reasoning LLM to understand the specific attributes of the MRI scans and the context of the diagnosis. Based on the provided context and MRI scan, the model reasoning the final diagnosis. The platform’s performance is evaluated across two key areas: Evaluation of Vision-Language Models and Evaluation of OpenAI-o3 Reasoning LLM.

\subsection{Diagnosis Evaluation of Vision Language Models}

\begin{figure}
\centering{}
\includegraphics[width=3.5in]{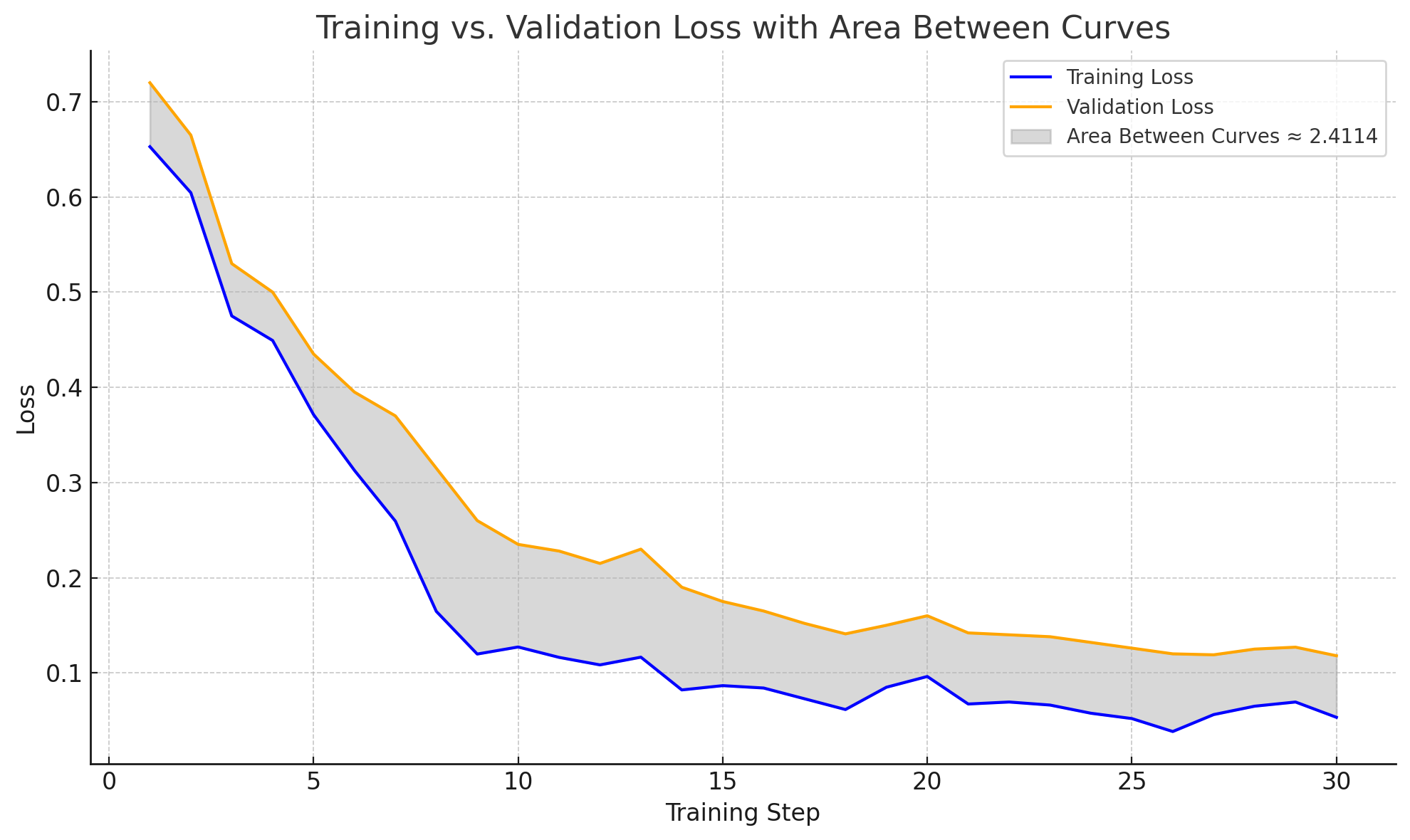}
\vspace{-0.1in}
\DeclareGraphicsExtensions.
\caption{Training loss and validation loss during fine-tuning of the Llama-3.2-11B-Vision-Instruct vision-language model.}
\label{unsloth-tranning-validation-loss}
\end{figure}

\begin{figure}
\centering{}
\includegraphics[width=3.5in]{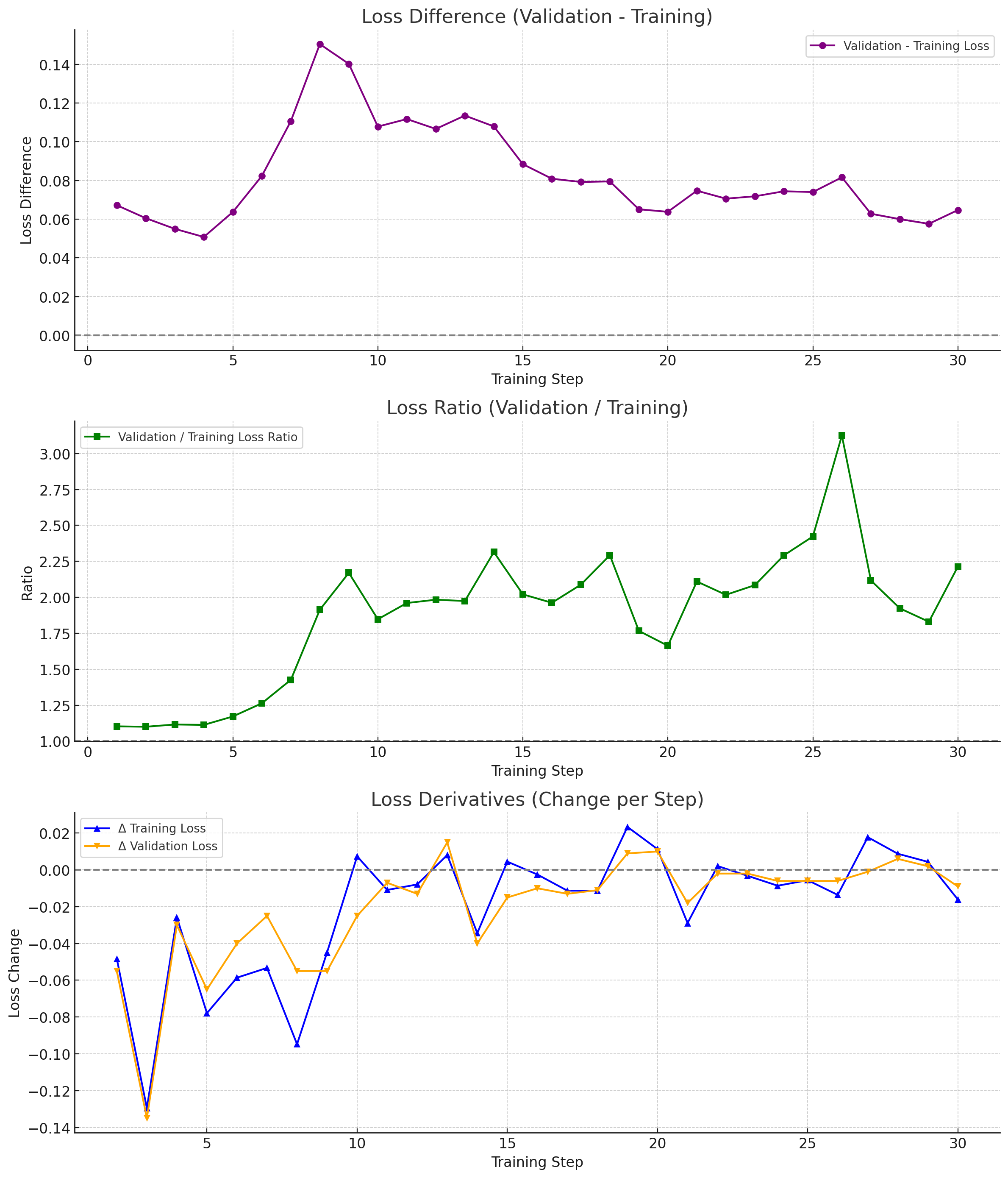}
\vspace{-0.1in}
\DeclareGraphicsExtensions.
\caption{Ratio of training to validation loss during the fine-tuning of the Llama-3.2-11B-Vision-Instruct vision-language model.}
\label{unsloth-loss-ratio}
\end{figure}

In this evaluation, we first measured the training and validation loss during the fine-tuning process of the vision-language models. These measurements, visualized in Figure~\ref{unsloth-tranning-validation-loss}, illustrate the model’s progressive improvement across training steps. Figure~\ref{unsloth-loss-ratio} further captures the rate of change in training and validation loss, providing insights into the convergence dynamics and stability of the models during fine-tuning.

\begin{figure}
\centering{}
\includegraphics[width=3.5in]{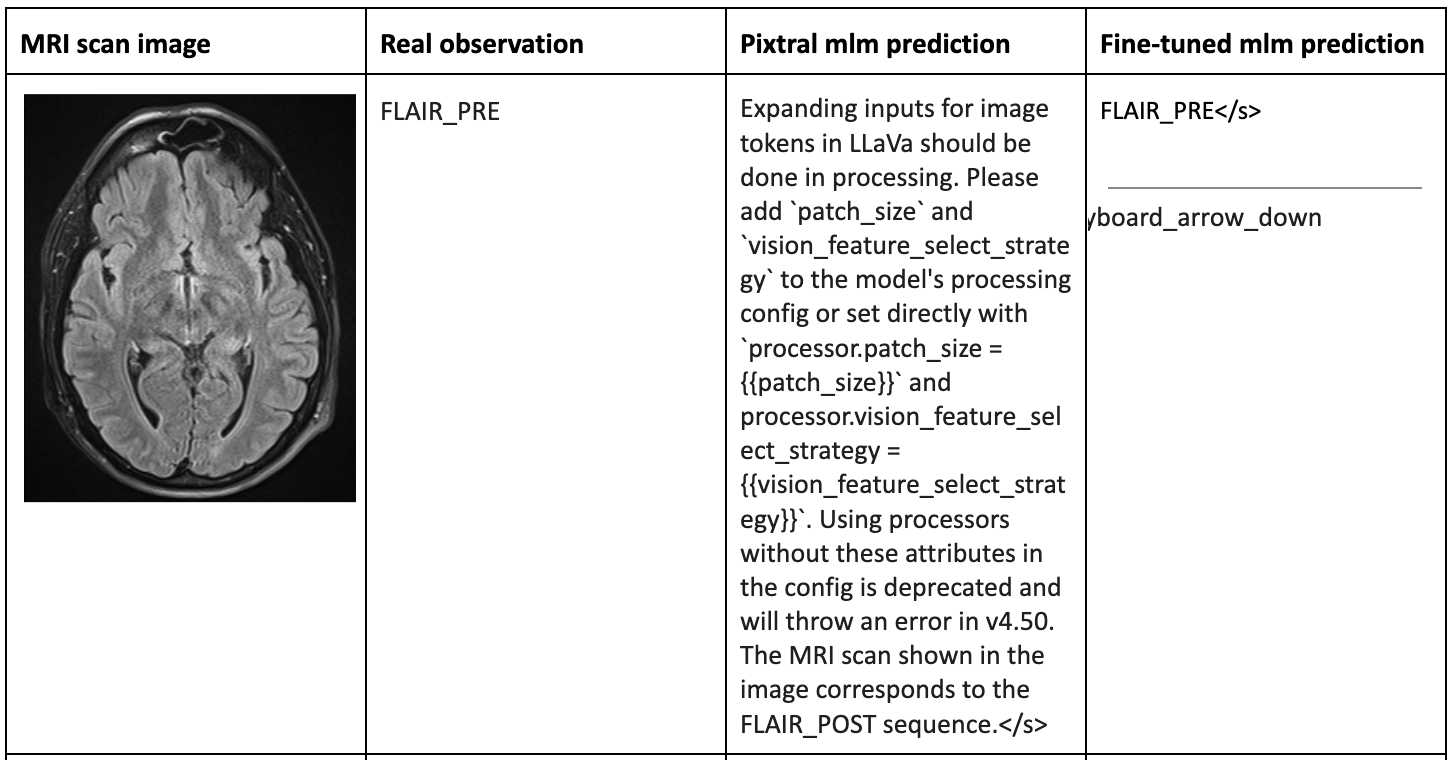}
\vspace{-0.1in}
\DeclareGraphicsExtensions.
\caption{The prediction results of Pixtral-12B-2409 vision language model.}
\label{z1}
\end{figure}

Next, we assessed the predictive performance of the fine-tuned vision-language models in diagnosing Traumatic Brain Injury (TBI) from MRI scans. This evaluation compared the real diagnostic observations of MRI images with predictions made by both the baseline vision-language models and their fine-tuned counterparts. Figure~\ref{z1} displays the output from the Pixtral-12B-2409 vision-language model, Figure~\ref{z2} shows the predictions from the Qwen2-VL-7B-Instruct model, and Figure~\ref{z3} illustrates the results from the Llama-3.2-11B-Vision-Instruct model.

The results clearly indicate that the fine-tuned models provide predictions that are significantly more precise and better aligned with expert diagnostic observations. Figure~\ref{confusion-metrics} presents the confusion matrix for the fine-tuned Llama-Vision model, illustrating its classification performance across four TBI-related MRI scan types. This visualization helps to identify which categories are most often confused, offering valuable insights for further model refinement.

To benchmark performance, we also trained a traditional image classification model using ResNet50~\cite{resnet-image-classification} on the same dataset. The outcomes of this model are shown in Figure~\ref{resnet}, providing a comparative baseline for evaluating the superiority of the fine-tuned vision-language models. Overall, these results demonstrate the effectiveness of the fine-tuning process and highlight the substantial improvements in diagnostic accuracy enabled by vision-language models in the context of TBI detection. 

\begin{figure}
\centering{}
\includegraphics[width=3.5in]{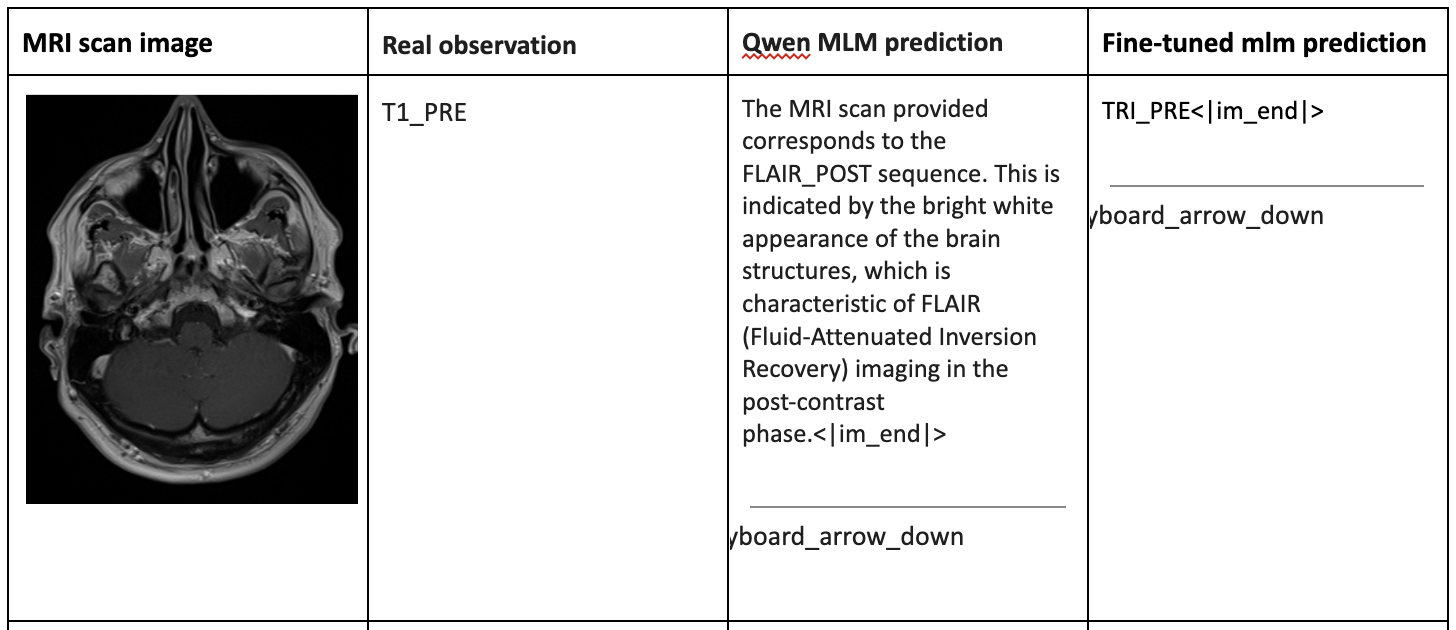}
\vspace{-0.1in}
\DeclareGraphicsExtensions.
\caption{The prediction results of Qwen2-VL-7B-Instruct vision language model.}
\label{z2}
\end{figure}

\begin{figure}
\centering{}
\includegraphics[width=3.5in]{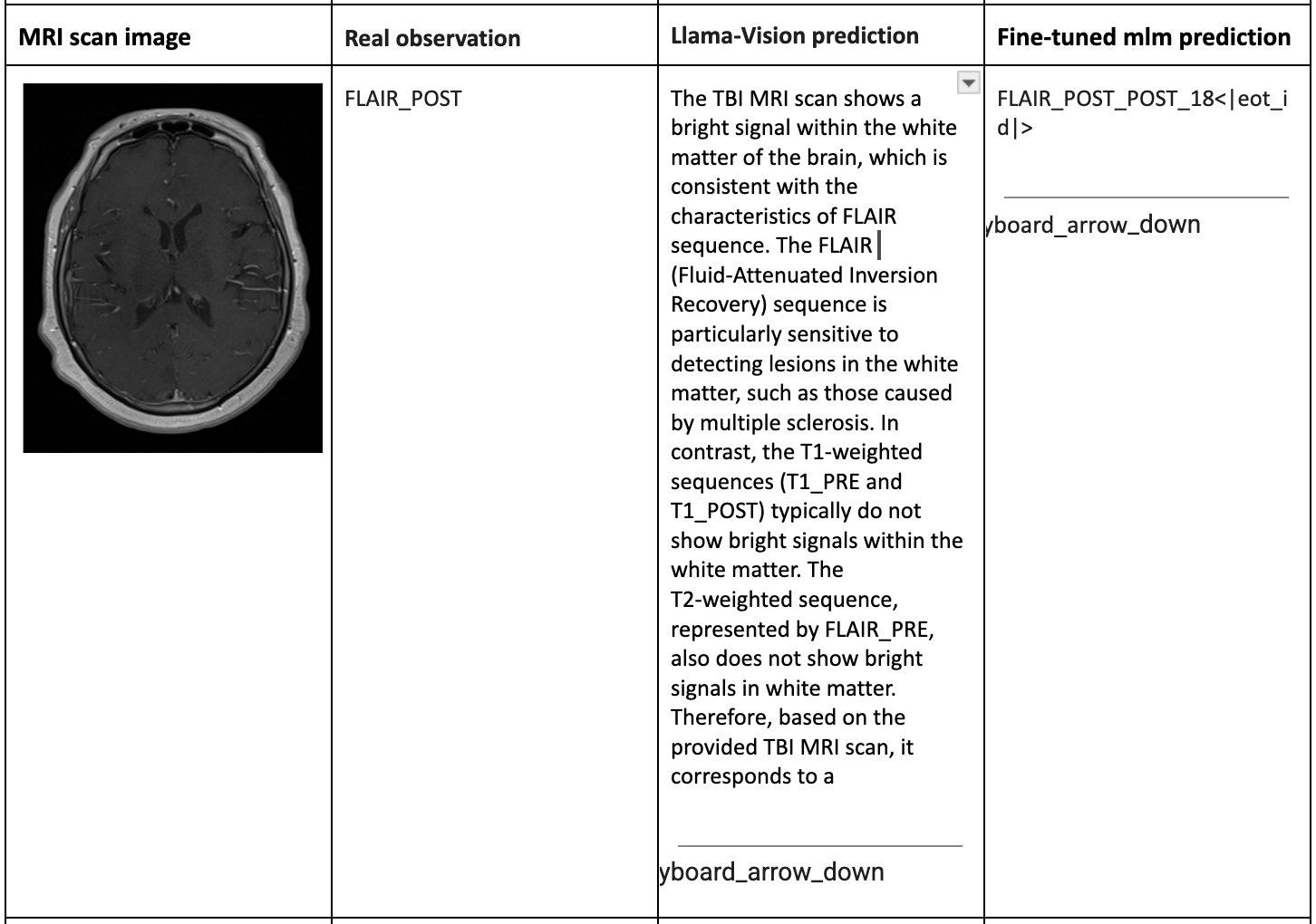}
\vspace{-0.1in}
\DeclareGraphicsExtensions.
\caption{The prediction results of Llama-3.2-11B-Vision-Instruct vision language model.}
\label{z2}
\end{figure}

\begin{figure}
\centering{}
\includegraphics[width=3.5in]{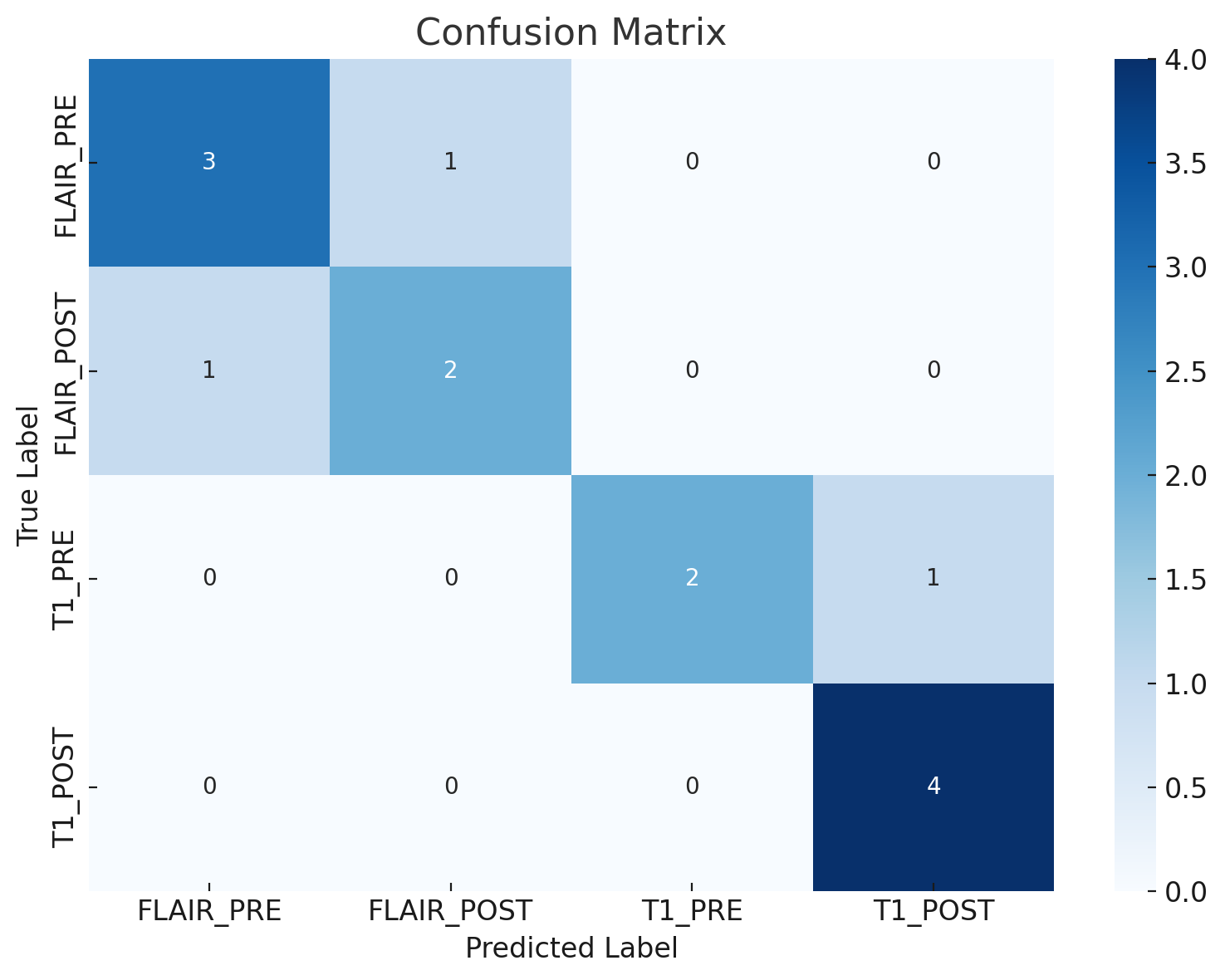}
\vspace{-0.1in}
\DeclareGraphicsExtensions.
\caption{Confusion matrix of the fine-tuned Llama-3.2-11B-Vision-Instruct vision-language model on TBI MRI scan classification.}
\label{confusion-metrics}
\end{figure}

\begin{figure}
\centering{}
\includegraphics[width=3.5in]{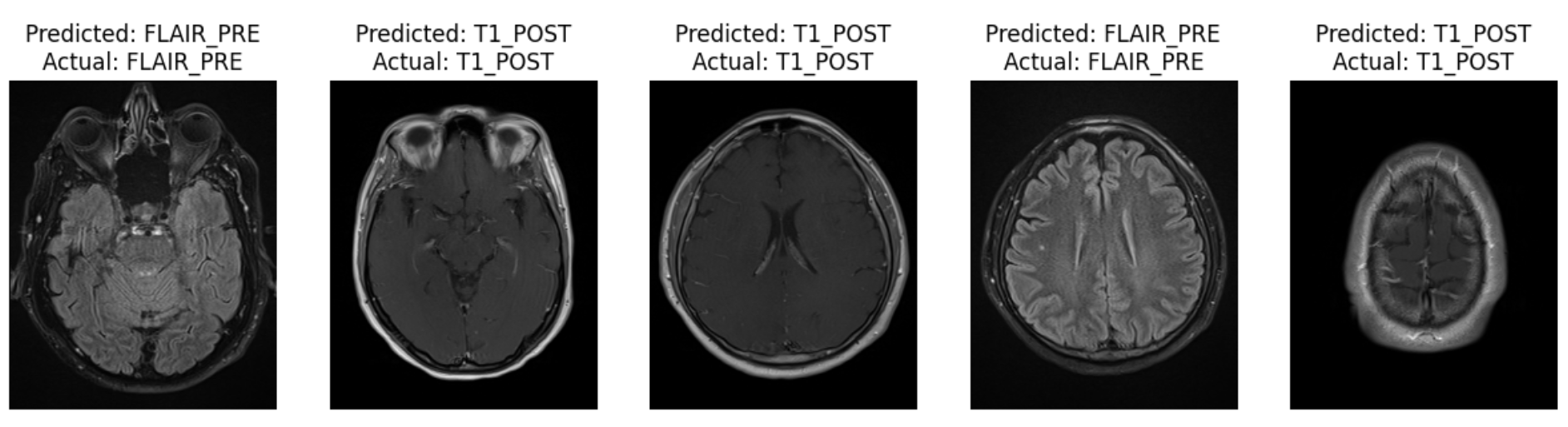}
\vspace{-0.1in}
\DeclareGraphicsExtensions.
\caption{Prediction results of the ResNet50 image classification model on the TBI MRI scan dataset.}
\label{resnet}
\end{figure}

\subsection{Diagnosis Reasoning Evaluation of the OpenAI-o3 LLM}

\begin{figure}[t]
\centering{}
\includegraphics[width=3.5in]{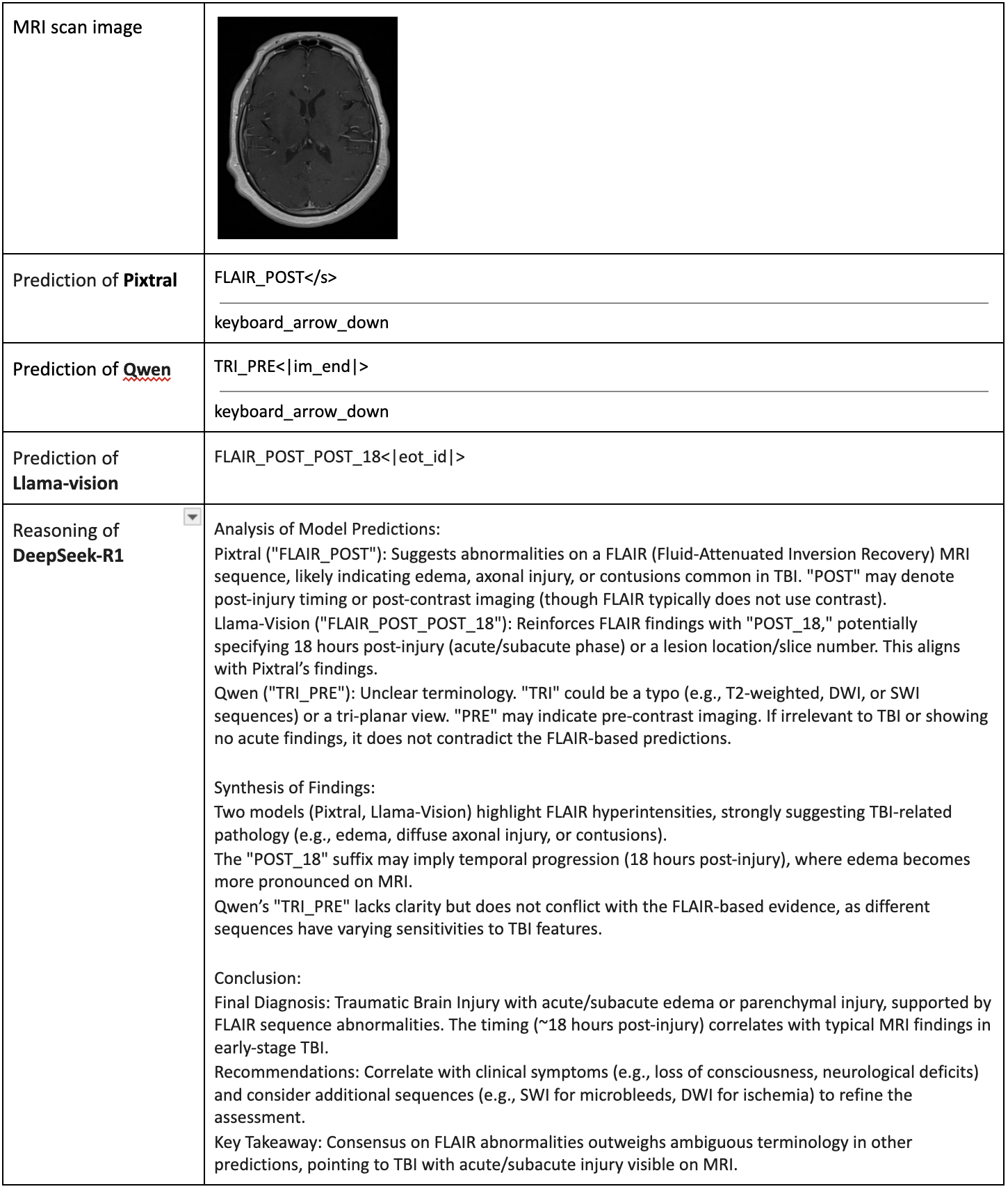}
\vspace{-0.1in}
\DeclareGraphicsExtensions.
\caption{Diagnosis reasoning made by OpenAI-o3 LLM.}
\label{z4}
\end{figure}

In this evaluation, we assessed the reasoning performance of the OpenAI-o3 Reasoning LLM by comparing the individual predictions from the vision-language models with the final diagnosis generated by OpenAI-o3. Figure~\ref{z4} illustrates the predictions made by different vision-language models for an MRI scan, along with the final diagnosis reasoning provided by OpenAI-o3. The results highlight OpenAI-o3’s ability to synthesize and analyze the outputs from multiple vision-language models, demonstrating its effectiveness and reliability in producing accurate final diagnoses. This evaluation underscores the reasoning LLM’s role in refining and enhancing diagnostic precision, further validating its integration into the Proof-of-TBI framework.

\begin{table*}[!htb]\centering
\vspace{0.1in}
\caption {LLM/VLM-based Medical Diagnosis Framework Comparison}
\begin{adjustbox}{width=1\textwidth}
\label{t_bc_platforms}
\begin{tabular}{lccccccc}
\toprule
\thead{Platform} & \thead{Domain} & \thead{Fine-tuning\\Support} & \thead{Running LLM} & \thead{Vision LM\\Support} & \thead{Reasoning LLM\\Support} & \thead{LLM Consortium\\Support} \\
\midrule
Deep-Psychiatrist & Psychiatric & \cmark & Llama-3, Pixtrel, Qwen2 & \cmark & \cmark & \cmark \\
Med-PaLM~\cite{med-palm} & General medicine & \cmark & PaLM & \xmark & \xmark & \xmark \\
LLM for DDx~\cite{llm-ddx} & General medicine & \cmark & Not specified & \xmark & \cmark & \xmark \\
Me-LLaMA~\cite{me-llama} & General medicine & \cmark & LLaMA & \xmark & \xmark & \xmark \\
CDSS~\cite{CDSS} & Metal Health & \xmark & GPT-4 & \xmark & \xmark & \xmark \\
DrHouse~\cite{drhouse} & General medicine & \cmark & Not specified & \cmark & \cmark & \xmark \\
Weda-GPT~\cite{wedagpt} & Indigenous Medicine & \cmark & Llama-3 & \xmark & \xmark & \xmark \\
\bottomrule
\end{tabular}
\end{adjustbox}
\end{table*}

\subsection{Related Works}

Several researchers have explored the use of large language models (LLMs) and vision-language models (VLMs) across various domains of healthcare diagnostics. Table~\ref{t_bc_platforms} presents a comparative analysis of these systems alongside our proposed Proof-of-TBI platform. Med-PaLM~\cite{med-palm}, developed by Google Research, is one of the earliest and most comprehensive medical LLMs, fine-tuned to answer clinical questions with high accuracy across general medicine. LLM for DDx~\cite{llm-ddx} focuses on generating differential diagnoses from clinical vignettes, demonstrating the potential of LLMs to assist clinicians in complex diagnostic reasoning. Me-LLaMA~\cite{me-llama} extends LLaMA models with continual pretraining and instruction tuning using biomedical literature and clinical notes, resulting in a strong foundation model tailored for general medical tasks. DrHouse~\cite{drhouse} introduces a multi-turn diagnostic reasoning system that integrates LLMs with sensor data and expert knowledge to simulate realistic patient-doctor interactions and improve diagnostic precision. CDSS~\cite{CDSS} proposes clinical decision support system for mental health diagnosis that combines LLMs and constraint logic programming. Weda-GPT~\cite{wedagpt} provide fine-tuned Llama-3 LLM-based decision support system for indigenous medicine.

\section{Conclusions and Future Work}

In this paper, we presented Proof-of-TBI, an innovative medical diagnosis support system that integrates a fine-tuned vision-language model consortium with the OpenAI-o3 Reasoning LLM to address the complex challenge of diagnosing mild Traumatic Brain Injury (TBI). By fine-tuning multiple vision-language models with labeled MRI scan datasets, we transformed general-purpose AI models into specialized tools capable of accurately identifying TBI-related symptoms. The system employs a consortium-based consensus mechanism to ensure reliability and accuracy in predictions, combining the insights of multiple models with the reasoning capabilities of the OpenAI-o3 LLM to deliver a final diagnosis. The OpenAI-o3 Reasoning LLM, which has demonstrated remarkable performance in reasoning tasks, analyzes the predictions from the fine-tuned vision-language models to produce the final TBI diagnosis from MRI scans. The LLM Agent orchestrates seamless interactions between the vision-language models and the OpenAI-o3 LLM, employing custom prompts to optimize communication and decision-making. The collaborative development of this platform with the U.S. Army Medical Research team in Newport News, Virginia, underscores its real-world applicability and transformative potential in the field of medical diagnostics. Evaluation results highlight the effectiveness of the platform, demonstrating its ability to deliver accurate, secure, and transparent TBI diagnoses. To the best of our knowledge, Proof-of-TBI is the first system to integrate fine-tuned vision-language models and reasoning LLMs for TBI prediction. This work paves the way for advancements in medical diagnostics, offering a scalable and secure framework that can be extended to other healthcare applications. By bridging AI and decentralized technologies, the platform establishes a novel paradigm for building reliable and trustworthy diagnostic systems. For future work, we intend to integrate additional open-source LLMs into the platform to further evaluate their performance and enhance its capabilities.

\bibliographystyle{unsrt}  
\bibliography{references} 

\begin{thebibliography}{10}

\bibitem{midl-tbi-definision}
Cl{\'e}mence Lefevre-Dognin, M{\'e}lanie Cogn{\'e}, Val{\'e}rie Perdrieau, Aur{\'e}lie Granger, Camille Heslot, and Philippe Azouvi.
\newblock Definition and epidemiology of mild traumatic brain injury.
\newblock {\em Neurochirurgie}, 67(3):218--221, 2021.

\bibitem{mild-tbi-challenges}
Ekaterina Lunkova, Guido~I Guberman, Alain Ptito, and Rajeet~Singh Saluja.
\newblock Noninvasive magnetic resonance imaging techniques in mild traumatic brain injury research and diagnosis.
\newblock {\em Human brain mapping}, 42(16):5477--5494, 2021.

\bibitem{tbi-ml-elsevior}
Maria~Jos{\'e} Uparela-Reyes, Lina~Mar{\'\i}a Villegas-Trujillo, Jorge Cespedes, Miguel Vel{\'a}squez-Vera, and Andr{\'e}s~M Rubiano.
\newblock Usefulness of artificial intelligence in traumatic brain injury: A bibliometric analysis and minireview.
\newblock {\em World Neurosurgery}, 2024.

\bibitem{vision-language-model}
Jingyi Zhang, Jiaxing Huang, Sheng Jin, and Shijian Lu.
\newblock Vision-language models for vision tasks: A survey.
\newblock {\em IEEE Transactions on Pattern Analysis and Machine Intelligence}, 2024.

\bibitem{vistion-language-model-comparison}
Jingyi Zhang, Jiaxing Huang, Sheng Jin, and Shijian Lu.
\newblock Vision-language models for vision tasks: A survey.
\newblock {\em IEEE Transactions on Pattern Analysis and Machine Intelligence}, 2024.

\bibitem{pixtral}
Pravesh Agrawal, Szymon Antoniak, Emma~Bou Hanna, Baptiste Bout, Devendra Chaplot, Jessica Chudnovsky, Diogo Costa, Baudouin De~Monicault, Saurabh Garg, Theophile Gervet, et~al.
\newblock Pixtral 12b.
\newblock {\em arXiv preprint arXiv:2410.07073}, 2024.

\bibitem{vlm-image-classification}
Fang Peng, Xiaoshan Yang, Linhui Xiao, Yaowei Wang, and Changsheng Xu.
\newblock Sgva-clip: Semantic-guided visual adapting of vision-language models for few-shot image classification.
\newblock {\em IEEE Transactions on Multimedia}, 26:3469--3480, 2023.

\bibitem{o3}
Gianluca Mondillo, Mariapia Masino, Simone Colosimo, Alessandra Perrotta, and Vittoria Frattolillo.
\newblock Evaluating ai reasoning models in pediatric medicine: A comparative analysis of o3-mini and o3-mini-high.
\newblock {\em medRxiv}, pages 2025--02, 2025.

\bibitem{reasoning-llms}
Yadong Zhang, Shaoguang Mao, Tao Ge, Xun Wang, Adrian de~Wynter, Yan Xia, Wenshan Wu, Ting Song, Man Lan, and Furu Wei.
\newblock Llm as a mastermind: A survey of strategic reasoning with large language models.
\newblock {\em arXiv preprint arXiv:2404.01230}, 2024.

\bibitem{qlora}
Tim Dettmers, Artidoro Pagnoni, Ari Holtzman, and Luke Zettlemoyer.
\newblock Qlora: Efficient finetuning of quantized llms.
\newblock {\em Advances in Neural Information Processing Systems}, 36, 2024.

\bibitem{ollama}
Tim Reason, Emma Benbow, Julia Langham, Andy Gimblett, Sven~L Klijn, and Bill Malcolm.
\newblock Artificial intelligence to automate network meta-analyses: Four case studies to evaluate the potential application of large language models.
\newblock {\em PharmacoEconomics-Open}, pages 1--16, 2024.

\bibitem{llamaindex}
Devin Schumacher.
\newblock V3ctron| data retrieval \& access system for flexible semantic search \& retrieval of proprietary document collections using natural language queries.
\newblock {\em Available at SSRN}, 2023.

\bibitem{langchain}
Simon Ott, Konstantin Hebenstreit, Valentin Li{\'e}vin, Christoffer~Egeberg Hother, Milad Moradi, Maximilian Mayrhauser, Robert Praas, Ole Winther, and Matthias Samwald.
\newblock Thoughtsource: A central hub for large language model reasoning data.
\newblock {\em arXiv preprint arXiv:2301.11596}, 2023.

\bibitem{tbi-mdpi}
Kuan-Chi Tu, Tee-Tau Eric~Nyam, Che-Chuan Wang, Nai-Ching Chen, Kuo-Tai Chen, Chia-Jung Chen, Chung-Feng Liu, and Jinn-Rung Kuo.
\newblock A computer-assisted system for early mortality risk prediction in patients with traumatic brain injury using artificial intelligence algorithms in emergency room triage.
\newblock {\em Brain sciences}, 12(5):612, 2022.

\bibitem{qwen2}
Peng Wang, Shuai Bai, Sinan Tan, Shijie Wang, Zhihao Fan, Jinze Bai, Keqin Chen, Xuejing Liu, Jialin Wang, Wenbin Ge, et~al.
\newblock Qwen2-vl: Enhancing vision-language model's perception of the world at any resolution.
\newblock {\em arXiv preprint arXiv:2409.12191}, 2024.

\bibitem{llamafactory-unsloth}
Yaowei Zheng, Richong Zhang, Junhao Zhang, Yanhan Ye, Zheyan Luo, Zhangchi Feng, and Yongqiang Ma.
\newblock Llamafactory: Unified efficient fine-tuning of 100+ language models.
\newblock {\em arXiv preprint arXiv:2403.13372}, 2024.

\bibitem{llama-3}
Abhimanyu Dubey, Abhinav Jauhri, Abhinav Pandey, Abhishek Kadian, Ahmad Al-Dahle, Aiesha Letman, Akhil Mathur, Alan Schelten, Amy Yang, Angela Fan, et~al.
\newblock The llama 3 herd of models.
\newblock {\em arXiv preprint arXiv:2407.21783}, 2024.

\bibitem{prompt-engineering}
Ggaliwango Marvin, Nakayiza Hellen, Daudi Jjingo, and Joyce Nakatumba-Nabende.
\newblock Prompt engineering in large language models.
\newblock In {\em International Conference on Data Intelligence and Cognitive Informatics}, pages 387--402. Springer, 2023.

\bibitem{resnet-image-classification}
Sheldon Mascarenhas and Mukul Agarwal.
\newblock A comparison between vgg16, vgg19 and resnet50 architecture frameworks for image classification.
\newblock In {\em 2021 International conference on disruptive technologies for multi-disciplinary research and applications (CENTCON)}, volume~1, pages 96--99. IEEE, 2021.

\bibitem{med-palm}
Karan Singhal, Shekoofeh Azizi, Tao Tu, S~Sara Mahdavi, Jason Wei, Hyung~Won Chung, Nathan Scales, Ajay Tanwani, Heather Cole-Lewis, Stephen Pfohl, et~al.
\newblock Large language models encode clinical knowledge.
\newblock {\em Nature}, 620(7972):172--180, 2023.

\bibitem{llm-ddx}
Daniel McDuff, Mike Schaekermann, Tao Tu, Anil Palepu, Amy Wang, Jake Garrison, Karan Singhal, Yash Sharma, Shekoofeh Azizi, Kavita Kulkarni, et~al.
\newblock Towards accurate differential diagnosis with large language models.
\newblock {\em arXiv preprint arXiv:2312.00164}, 2023.

\bibitem{me-llama}
Qianqian Xie, Qingyu Chen, Aokun Chen, Cheng Peng, Yan Hu, Fongci Lin, Xueqing Peng, Jimin Huang, Jeffrey Zhang, Vipina Keloth, et~al.
\newblock Me-llama: Medical foundation large language models for comprehensive text analysis and beyond.
\newblock 2024.

\bibitem{CDSS}
Brian~Hyeongseok Kim and Chao Wang.
\newblock Large language models for interpretable mental health diagnosis.
\newblock {\em arXiv preprint arXiv:2501.07653}, 2025.

\bibitem{drhouse}
Bufang Yang, Siyang Jiang, Lilin Xu, Kaiwei Liu, Hai Li, Guoliang Xing, Hongkai Chen, Xiaofan Jiang, and Zhenyu Yan.
\newblock Drhouse: An llm-empowered diagnostic reasoning system through harnessing outcomes from sensor data and expert knowledge.
\newblock {\em Proceedings of the ACM on Interactive, Mobile, Wearable and Ubiquitous Technologies}, 8(4):1--29, 2024.

\bibitem{wedagpt}
Eranga Bandara, Peter Foytik, Sachin Shetty, Ravi Mukkamala, Abdul Rahman, Xueping Liang, Ng~Wee Keong, and Kasun De~Zoysa.
\newblock Wedagpt—generative-ai (with custom-trained meta’s llama2 llm), blockchain, self sovereign identity, nft and model card enabled indigenous medicine platform.
\newblock In {\em 2024 IEEE Symposium on Computers and Communications (ISCC)}, pages 1--6. IEEE, 2024.

\end{thebibliography}

\end{document}